# Incremental Learning Techniques for Online Human Activity Recognition

Meysam Vakili[1*], Masoumeh Rezaei[2]

1 Department of Computer Engineering, University of Science and Culture, Tehran, Iran
2 Department of Computer Engineering, Khorasan Institute of Higher Education, Razavi Khorasan, Iran
* Corresponding author: m.vakili@usc.ac.ir;

**Abstract**

**Unobtrusive and smart recognition of human activities using smartphones inertial sensors is an interesting topic in the field of artificial intelligence acquired tremendous popularity among researchers, especially in recent years. A considerable challenge that needs more attention is the real-time detection of physical activities, since for many real-world applications such as health monitoring and elderly care, it is required to recognize users' activities immediately to prevent severe damages to individuals' wellness. In this paper, we propose a human activity recognition (HAR) approach for the online prediction of physical movements, benefiting from the capabilities of incremental learning algorithms. We develop a HAR system containing monitoring software and a mobile application that collects accelerometer and gyroscope data and send them to a remote server via the Internet for classification and recognition operations. Six incremental learning algorithms are employed and evaluated in this work and compared with several batch learning algorithms commonly used for developing offline HAR systems. The Final results indicated that considering all performance evaluation metrics, Incremental K-Nearest Neighbors and Incremental Naïve Bayesian outperformed other algorithms, exceeding a recognition accuracy of 95% in real-time.**

## 1. Introduction

During the last years, there has been an exponential growth in the development of smartphones which can support a broad range of applications. Simultaneously, the evolution of mobile and pervasive computing and the emergence of state-of-the-art AI techniques have pushed the boundaries of smartphones capabilities enormously. One of the most fascinating applications acquired immense popularity among both scientists and industrialists is the automatic detection of human activities using various technologies such as sensors and cameras, which is usually called Human Activity Recognition or HAR. Today, HAR plays a pivotal role in many domains, such as healthcare [1, 2], transportation [3, 4], security [5, 6], etc.

Considering data acquisition technologies and devices, there are two major types of activity recognition approaches [7, 8] proposed in the literature: 1) vision-based activity recognition [9], which performs activity detection process through videos or images captured by cameras, and 2) sensor-based activity recognition [10] which leverages different types of inertial sensors such as accelerometers, gyroscopes and magnetometers in order to distinguish various physical activities. Due to drawbacks and limitations of vision-based systems including expensive equipment, heavy reliance on lighting condition and privacy issues [11-14], sensor-based approaches have achieved greater attention and have obtained superior performance [15].

Nowadays, most of the modern smartphones benefit from a considerable number of built-in sensors which among them are inertial sensors. These sensors that are controlled by mobile APIs are capable of providing similar functions to wearable sensors for HAR, while remaining unobtrusive and do not make smartphone owners feel inconvenient [16]. In addition, the ever-increasing computational power and storage space of these devices have enabled them to perform complicated calculations and run complex machine and deep learning algorithms. Thus, all of these features have made todays' smartphones major platforms for human activity recognition and have dramatically raised their prevalence in assistive technology research [17, 18].

HAR problems are basically considered as conventional pattern recognition problems, and more precisely, classification problems [19]. Owing to this, most of the proposed human activity recognition solutions are based upon AI techniques which benefit from machine and deep learning algorithms [20, 21]. According to research studies, there are two general types of human activity recognition systems [15, 22]: offline HAR, which is much more popular and most of the research work fall into this category, and online HAR. In Offline HAR, traditional batch learning methods are used, and all phases of activity recognition including data collection, data annotation (or labeling), data preprocessing, feature extraction and classification are done offline. It means, all required data for making a robust model are at hand from the beginning and classification algorithms are trained on the entire dataset and are used for prediction thereafter. Thus, offline HAR approaches consist of fixed learning models that disregards the dynamic changes of human behavior and do not consider the fact that various people perform similar activities in different styles [23]. Once new input data from new subjects arrive, if these models want to update themselves, it is required to retrain the whole model from scratch, which is a massive waste of time and resources. According to the characteristics of offline HAR solutions, it is clear that they are not suitable for real-world applications such as health monitoring, fall detection and people motion tracking systems that need real-time detection of users' activities.

As a result, there is no option but to employ online HAR. This approach makes use of online learning techniques which are able to construct models incrementally from a sequence of data arriving continuously [24, 25] and because of this, online machine learning also called incremental learning. Contrary to batch learning algorithms, incremental learning algorithms can update existing models immediately in the arrival of new training data instances, leading to reduced training time and memory usage, in addition to the efficient handling of real-time changes in data distribution [7]. Thus, online HAR approaches are much more robust and scalable for real-world applications where not only are data massive in size, but also have high velocity.

It is well worth mentioning that, by online or real-time machine learning, we mean our incremental algorithms can build models in real-time using training data, and real-time testing (prediction) is not of great importance. Because it is clear that once a model has been trained, it is ready to predict target labels for new input instances in almost real-time.

Furthermore, HAR approaches can be general or personalized, also called user-independent and user-specific, respectively [26]. Classification models in general methods are built using datasets collected from a number of users, and these models are employed for activity detection of new users who are different from primary subjects. The major challenge is that people are different regarding their personal attributes, namely physical characteristics, health conditions, gender, etc. [27, 28]. Thus, even if the main dataset is huge and comprises millions of activity data from a countless number of users, it is still quite possible that there are other people having various activity patterns and do the same activity in a different fashion [29]. To address this issue, it is required to design personalized HAR systems which are able to generate user-specific models by means of each subject's personal data. Due to the features and capabilities of incremental learning techniques to update and adapt themselves with newly arrived data, they can be leveraged to create effective personalized HAR models [27, 29, 30].

According to our extensive studies, in comparison with offline HAR methods, online approaches have received much less attention. Moreover, studies related to using incremental learning algorithms for recognizing physical activities are still in their infancy and much more research work is required to be conducted. In fact, proposed work are not comprehensive and most of them face challenges which will be discussed in more detail in the next section. Thus, in this paper, we are going to deeply investigate the procedure of designing a human activity recognizer using online machine learning, in order to propose an approach for detecting physical activities in real-time. In fact, we leverage the incremental version of several popular machine learning algorithms to examine their capabilities in the creation of an online recognizer. Considered algorithms are Incremental K-Nearest Neighbors, Incremental Decision Trees, Incremental Random Forests, Incremental Adaboost, Incremental Naïve Bayesian and Learn++NSE. Due to the online nature of our approach, it has the ability to adapt to new users' activity patterns easily, and therefore, it is also suitable for personalized activity recognition. In order to make our solution much closer to real-world situations, instead of using existing datasets, we establish our own data collection environment by developing a specific mobile application. We use smartphone inertial sensors including accelerometer and gyroscope as the source of input data streams. It is also noticeable that rather than just a few common activities, we use a much wider range of activities for detection process. Presented algorithms are assessed based on different performance evaluation metrics. Moreover, we compare our incremental algorithms with their offline counterparts as baselines.

Ultimately, using incremental learning algorithms producing the best performance in our experiments, we develop an online HAR software which is able to work well with our mobile data collector application via the Internet.

The rest of the paper is organized as follows. Section 2 discusses related works that used online machine learning and inertial sensors to propose HAR systems. In section 3, the methodology is described in great details. Section 4 evaluates the presented approach and algorithms and compares them with some batch learning methods. Finally, section 5 concludes this paper with future work.

## 2. Background

In this section, we first provide a brief introduction to incremental learning and where it comes from. Afterwards, we carry out a literature review on research works which leveraged online or incremental learning to propose HAR systems. All of them benefited from inertial sensors data collected either by smartphones or wearables.

### 2.1. Incremental Learning

Traditional machine learning or batch learning methods where the entire dataset is accessible at the beginning cannot meet the requirements of handling the huge volume of data in real-time [31]. Moreover, they cannot integrate new information into already built models continuously and should reconstruct new models from scratch in the presence of new data samples. Clearly, not only is this process extremely time-consuming, but it also results in creating obsolete models.

To overcome these issues, we need to take advantage of sequential data processing. In this way, we can make use of data as soon as it is available, resulting in having models which are up to date all the time, while cutting down costs of maintenance and data storage. Incremental learning algorithms perfectly suit this scheme, in that they continuously modify created models with new input data and minimize processing time and storage space. The remarkable abilities of incremental learning algorithms in continuous real-time and large-scale processing have led to gaining more attention, especially in the context of Big Data [32].

However, there are still some ambiguities in defining incremental and online learning in the literature. Some researchers use them interchangeably, whereas others differentiate them in various ways [7]. Lifelong or continual learning are also additional terms used by some authors. We are not going to deeply investigate the details about incremental machine learning here, but just as a straightforward and brief description according to Losing et.al [33] *"an incremental learning algorithm generates a sequence of models using a given stream of training data, where each model is only dependent on the previous one and very few recent input samples"*. Thus, incremental learning models can be constructed without complete retraining, while being able to preserve previously obtained knowledge from prior models. More information about this type of learning can be found in [24, 32, 33].

### 2.2. Related Work

In [7], a class incremental learning approach for activity recognition was proposed, which can detect new classes of activities continuously. The paper presented a new algorithm based on traditional Random Forests called Class Incremental Random Forests (CIRF). In fact, it combines traditional and its new separating axis theorem (SAT) based splitting strategies in order to reduce information loss and control the growth of trees. Three public datasets were used for evaluation, and the performance of the CIRF was compared with some similar methods. [34] introduced a HAR system based on a popular incremental learning algorithm, Learn++. It leverages an already available dataset of human activities alongside a feedback correction mechanism receiving true labels from users in case of false predictions occurred by the system. In this work, wearable accelerometer and gyroscope sensors were used in three different body positions, naming chest, wrist and ankle. Zhao et al. [35] proposed a class incremental version of batch Extreme Learning Machine (CIELM) for predicting human physical behavior. They extended online sequential extreme learning machine (OSELM) [36] to enable their solution to learn and recognize new activities incrementally. In fact, they first trained an ELM classifier on a set of collected data and then exploited OSELM and new classes of data to adapt the old ELM to a new classifier gradually. For data collection, the accelerometer sensor of a smartphone was used while it was positioned in the subject's trouser pocket. Furthermore, the authors applied CIELM on two existing image datasets to demonstrate the suitability of their method for other application domains.

Authors in [37] presented a personalized and adaptive framework for activity recognition, which is in fact, an improvement of their previous work, StreamAR [38]. It consists of three main components including modeling, adaptation and recognition components. The adaptation component comprises incremental and active learning, while an ensemble classifier has been employed in the recognition component. Three public datasets were used for evaluation, and experiments were done on both a smartphone and a PC. An unsupervised clustering approach for activity recognition was presented in [16], which is based upon traditional K-Means and consists of 2 stages. In the first stage, accelerometer and gyroscope data which come from mobile phone sensors are processed one by one using an improved version of single-pass streaming clustering algorithm proposed in [39]. The output of this stage is then sent to the second stage, where pre-clustered data are modified and final clusters of activities are created incrementally. A dedicated mobile app was designed to perform all phases of activity recognition process. In [40], an incremental learning approach for activity recognition was proposed, which employs an ensemble of Support Vector Machines (SVMs). It uses the first batch of data to train a model. Afterwards, it discards the data classified correctly and trains a new model on the data samples classified incorrectly. Training new models on subsequent batches of data will continue to produce an SVM for each misclassified batch. For obtaining the final classifier, weighted sums of all models trained correctly and the models trained on the misclassified samples are integrated. Weight assignments process is performed using Particle Swam Optimization. A public dataset was used for evaluation and the performance results of the presented method and normal SVM was compared.

The work in [41] presented MOARS (Mobile Online Activity Recognition System) with the aim of providing an online HAR framework which is resistant to the changes of a smartphone's orientation. For solving this issue, the authors proposed to make use of the Earth coordinate system instead of a smartphone coordinate system and to further use magnetometer and gyroscope in association with accelerometer. Four common machine learning algorithms including RF, NB, KNN and SVM were exploited in MOARS, and for online training and prediction, it benefits from real-time users feedback receiving by a dedicated mobile application. [42] investigated the capabilities of three machine learning algorithms naming KNN, C4.5 and RIPPER in identification of people's movements in both offline and incremental manner. However, in practice, authors only implemented the incremental version of KNN, and for other algorithms, whenever a new data sample arrived, they retrained the whole model. They also developed a mobile application for collecting accelerometer data and just used three features and a frequency of 1Hz to reduce the resource consumption of smartphones. The mobile device's position for data collection was user's pocket (it was not mentioned which pocket).

Siirtola et al. [30] proposed an incremental learning method for personalized HAR. In particular, they leveraged ensemble Learn++ algorithm with three different base learners, including LDA, QDA and CART. Their approach comprises two phases. First, user-independent data are collected to train some base classifiers, which are then added to Learn++ model. The second phase starts when the application (implemented in a smartphone or another smart gadget) is used by a user, and personal streaming data are obtained for updating and personalizing the model. All three types of learning, namely supervised, semi-supervised and unsupervised were employed for personalized detection, in order to label classes of new activities. Experiments were conducted using three public datasets. A subject-specific activity recognition solution using incremental SVM was proposed in [43], that is in fact, an extension of the authors' previous method in [44]. They first build a non-personalized or general model using some training data and then update it with new input samples of a specific user to produce a personalized model. For updating process, active learning was exploited to receive true labels from users. An uncertainty evaluation is done in online phase and users are only requested to label data in which the uncertainties of predictions are under a predefined threshold. Streaming data were gathered by a wrist-worn accelerometer for detecting a wide range of activities in four categories, including ambulation, cycling, sedentary and other.

Reviewing some major related projects, we can see that we are faced with some challenges. One significant challenge is that many of them have used existing datasets. For offline HAR, it is acceptable that researchers employ publicly available datasets. But for online HAR, we believe that real-time data streams should be used to provide the proposed approach with real-world situations. In the real world, data is produced in real-time and our knowledge about input data is highly restricted. We do not know too much about coming data beforehand and we are not aware of patterns followed by input data samples. Moreover, the data can be noisy, while some of the existing datasets have been already cleaned by different preprocessing tasks. Thus, if we use public datasets, we do not really simulate real-world conditions. Another challenge is that, many of the presented works have used a training stage, though small, before real-time predictions. It is clear that this pre-training phase can boost the prediction power of their solutions, but it makes the system far from a truly real-time HAR system which should start activity recognition process from the beginning, without being trained already. Also, not only did most of these papers employ very few common physical activities, but they also never explored a wide range of online machine learning methods. In this work, we aim to address these challenges and more deeply investigate the design of an online HAR solution by means of incremental learning algorithms.

## 3. Online Human Activity Recognition

In this section, we are going to describe our approach and analyze our HAR system in more detail. As mentioned earlier, the primary purpose of this research work is to explore the suitability of incremental learning algorithms in developing an activity recognition solution and then propose a framework for online HAR using top-performing algorithms. It should be noted that at this stage, we do not aim to develop a thorough and elaborate HAR product for market use. In fact, the presented system is just a simple prototype of our HAR approach for probable future development.

In general, the system has a client-server architecture containing three main modules: data collection module on the client side and preprocessing and recognition modules on the server side. Activity data are captured by smartphones accelerometer and gyroscope sensors at the user's location and sent to the preprocessing module on the server through the Internet. Finally, the recognition module creates a base model using incremental classification algorithms, and the activity detection process is performed by this module. To be more exact, an initial model is created and true labels (annotated target labels) are then injected into the system via the user's feedback. We developed a mobile application for data collection on the client side and dedicated computer software on the server side to monitor, manage and display the activity recognition process. More details about the operations of each module will be presented in the following subsections.

### 3.1. Data Collection

To collect required data, instead of using already-prepared datasets, that is quite popular method in HAR studies, we generate activity data through real-time streams produced by smartphone sensors. As we discussed in section 2, this can help make our work much closer to real-world situations. Data collection is done using a Xiaomi Redmi k20 smartphone. We use accelerometer and gyroscope sensors to generate continuous streams of input data for the recognition system. Four subjects, including two men and two women, participated in our data collection phase.

Almost all research works in the field of activity recognition detect a limited number of predefined activities, such as sitting, standing, walking, running, etc. But in our work, we eliminate this limitation to a great extent and give users more freedom to do a much wider range of activities. Therefore, the list of our activities, shown in table 1, consists of some typical activities, plus a number of light physical exercise activities. Detection of these activities in real-time and with high accuracy will make our proposed solution entirely suitable for some applications in the realms of physical therapy and remote coaching. To collect data, we place the mobile device in either of these two positions: inside the trousers' pocket or on the forearm using a cell phone armband. The activities are separated in terms of the smartphone position in table 1.

**Table 1**. Physical activities based on the smartphone position

| No. | Activity (inside trouser)      | No. | Activity (on forearm)  |
|-----|--------------------------------|-----|------------------------|
| 1   | Walking                        | 13  | Arm Swings             |
| 2   | Running                        | 14  | Forearm Rotation       |
| 3   | Standing Still                 | 15  | Dumbbell Biceps Curl   |
| 4   | Sitting on a Chair             | 16  | Jumping Jack           |
| 5   | Side Leg Lifts                 | 17  | Chest Expansion        |
| 6   | Boxer Shuffle                  | 18  | Cross Toe Touch        |
| 7   | Knee Lifts                     | 19  | Straight Punch         |
| 8   | Cycling using Exercise Bicycle | 20  | Big Arm Circles        |
| 9   | Forward Lunge                  |     |                        |
| 10  | Torso Rotation                 |     |                        |
| 11  | Squats                         |     |                        |
| 12  | Mountain Climber Twist         |     |                        |

To facilitate sensor data gathering, management and transition, we have developed our dedicated Android application (Fig. 1). Defining new activities, changing data collection frequency and storing sensors data as CSV files are other features of this application. Users can define any number of activities in the application. When they decide to perform an activity, they should first annotate the data by specifying the type of activity. Afterwards, by pressing the sense button, accelerometer and gyroscope data along with the corresponding labels are sent to the server via the Internet, so that detection operation can be performed there. Employing users' feedback to identify true target labels is a crucial technique in HAR in order to increase the accuracy of the system. This method, also called active learning in some literature [38, 45], becomes much more important in online HAR. The reason is that when data streams are first entered into the system, the system has no prior or background knowledge about the type of activities being performed. It is quite similar to "cold start" problem in recommendation systems [46]. Hence, the need for true labels to train and update underlying classifiers is an essential requirement of online HAR, which results in boosting the recognition accuracy of the system for future predictions.

## 3.2. Preprocessing

The preprocessing phase is one of the inseparable parts of almost all HAR systems that have been designed so far. However, the type and number of preprocessing tasks done in online and offline HAR systems can be quite different. In offline HAR, there is no time restriction in performing activity detection operations; therefore, a wide variety of preprocessing techniques can be applied before running classification algorithms. As a consequence, offline systems would be able to obtain excellent results. But in online systems, a significant challenge is response or recognition time, which is the interval between the time a user performs an activity and the time classification algorithm detects that activity. Undoubtedly, one of the major goals of online HAR is to reduce response time; thus, some of the preprocessing operations cannot be performed so as to minimize overall workload and response time. Especially when a HAR system is fully implemented on portable devices such as smartphones or smart watches, reducing computations and preprocessing tasks become enormously important, since they are much more limited in terms of energy and computing resources than powerful computers.

On the other hand, preprocessing tasks cannot be eliminated completely, because it leads to a significant decrease in the overall performance of the system. Due to these issues, there is a challenging trade-off between response time and the recognition accuracy of an Online HAR approach. In general, preprocessing phase can involve various operations, but the most important ones in activity recognition problems are data normalization and feature extraction. In our method, a module is allocated to preprocessing phase including components for normalizing data and extracting features. Employing these data preparation tasks, particularly feature extraction, considerably improved our final results. Feature extraction has always been an integral part of preprocessing phase in any HAR system, which is performed on raw data before applying classification algorithms. It is worth mentioning that in some research projects benefiting from neural network solutions, this step is automatically removed. The reason is that neural networks such as CNNs use their unique techniques to extract a number of complex features that are different from conventional and well-known features commonly used in HAR systems. However, these algorithms and their implementation details are not the subjects of this paper and interested readers in NN-based HAR can look at studies conducted in [47, 11].

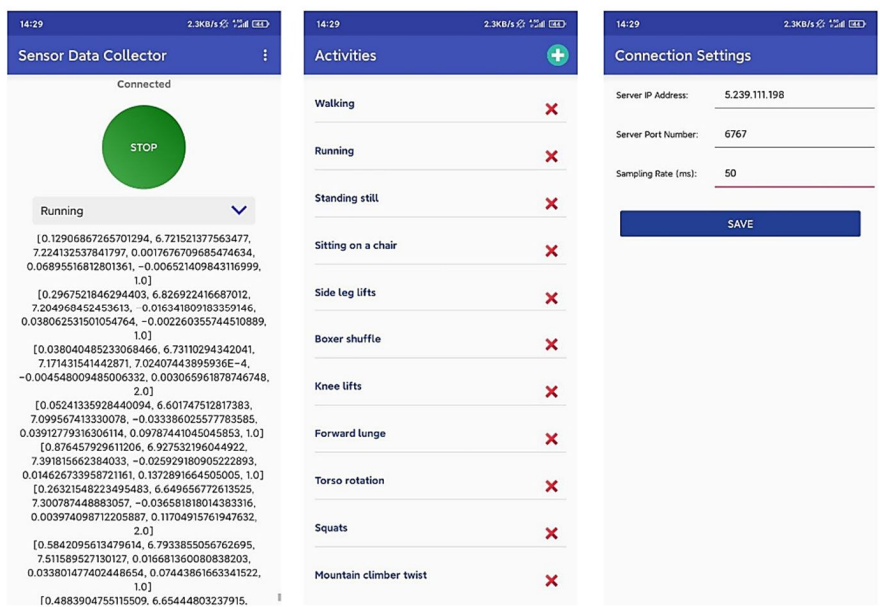

**Fig. 1**. Data collector application

Table 2 shows some of the most widespread time-domain and frequency-domain features [7, 19] we extracted from raw sensor data. For the precise and reliable implementation of these features, in addition to Python's scientific libraries, two high level and domain-specific libraries named tsfresh [48] and tsfel [49] have been employed.

So far, many studies have been conducted to determine the best frequency for data collection in HAR systems [50-52]. According to the thorough experiments of Khusainov et al. [53], a frequency of 20 Hz is totally sufficient for recording one complete human activity with required details. Therefore, we use this frequency, which means 20 data samples are collected from the sensors every second. Sliding window (*sw*) size is 2 seconds and with no overlapping. Therefore, each window includes 40 raw sensor data samples. The mentioned features are then extracted from each window in order to create a unique feature vector with size 98 (for each sensor, 16 features for each axis and a Signal Magnitude Area feature for a synthesis of all three axes). This vector is then sent to the recognition module for model generation.

Table 2. Extracted features from raw sensors data

| Feature | Formula | Extra Description |
|---|---|---|
| **Maximum** | $max(sw_i)$ | $sw = (s_1, s_2, \dots, s_n)$ |
| **Minimum** | $min(sw_i)$ | |
| **Mean** | $\mu_s = \frac{1}{n}\sum_{i=1}^{n} s_i$ | |
| **Median** | $median(sw_i)$ | |
| **Standard Deviation** | $\sigma_s = \sqrt{\frac{1}{n}\sum_{i=1}^{n}(s_i - \mu_s)^2}$ | |
| **Range** | $max(sw_i) - min(sw_i)$ | |
| **Skewness** | $\frac{1}{n\sigma_s^3}\sum_{i=1}^{n}(s_i - \mu_s)^3$ | |
| **Kurtosis** | $\frac{1}{n\sigma_s^4}\sum_{i=1}^{n}(s_i - \mu_s)^4$ | |
| **Interquartile Range** | $quartile3(sw_i) - quartile1(sw_i)$ | |
| **Autocorrelation** | $\frac{1}{(n-k)\sigma_s^2}\sum_{i=1}^{n-k}(s_i - \mu_s)(s_{i+k} - \mu_s)$ | |
| **Root Mean Square** | $\sqrt{\frac{1}{n}\sum_{i=1}^{n} s_i^2}$ | |
| **Maximum Frequency** | $maxFreq(sw_i)$ | |
| **Median Frequency** | $medFreq(sw_i)$ | |
| **Spectral Centroid** | $\frac{\sum_{i=1}^{n} k(i)f(i)}{\sum_{i=1}^{n} f(i)}$ | $f(i)$ = weighted frequency value of bin *i*<br>$k(i)$ = center frequency of bin *i* |
| **Spectral Entropy** | $-\sum_{i=1}^{n} p_i \log p_i$ | $p_i = \frac{s_i^2}{\sum_{j=1}^{n} s_j^2}$ |
| **Spectral Energy** | $\frac{1}{n}\sum_{i=1}^{n} s_i^2$ | |
| **Signal Magnitude Area** | $\frac{1}{n}\sum_{i=1}^{n}(|x_i| + |y_i| + |z_i|)$ | |

## 3.3. Activity Recognition

To recognize users' physical activities in real-time or close to real-time, we need models that can learn as fast as possible and with very few input training samples. To achieve this goal, we benefit from online machine learning methods, or incremental learning algorithms, to be exact. The algorithms we use in this work are, in fact, incremental versions of some of the most widely used traditional classification algorithms, including Incremental K-Nearest Neighbors (IKNN), Incremental Decision Trees (IDT), Incremental Random Forests (IRF), Incremental AdaBoost (IAdaBoost) and Incremental Naïve Bayes (INB). We also use Learn++NSE [54], the non-stationary environment version of Learn++ [55], which is a well-known incremental learning algorithm. According to our extensive explorations, just a few research projects have leveraged incremental learning to address the issue of online HAR, and most of them have focused on one specific algorithm. One of our main contributions in this study is to use several incremental learning algorithms in order to make a comprehensive evaluation and compare their performances. All of the algorithms used in this work have been implemented based on credible sources [56-58] and with the use of two powerful libraries called Creme [59] and Scikit-Multiflow [60] developed particularly for stream processing and online learning.

In addition to the main contributions of our work which are providing an approach for online human activity detection and investigating the suitability of incremental learning algorithms to do this, we have developed a client-server based HAR system to make all procedures completely clear and tangible. In fact, after conducting extensive experiments, which we will discuss in complete details in the next section, we designed an online HAR framework using our best-performing incremental learning algorithms (IKNN and INB). Although this system has been designed to be relatively simple at this stage, it is quite capable of becoming a comprehensive system for online activity recognition, or at a more advanced level, for IoT-based HAR. To the best of our knowledge, most of the proposed HAR systems are more research and laboratory-based, and therefore, all activity recognition operations take place within a local network. However, our system has been basically designed to use the Internet as the main platform for data transfer. We have used Python WebSockets in order to facilitate and speed up sensor data transmission over the Internet. We have also designed a HAR Monitoring Software with a simple UI on the server side which is shown in Fig. 2. Using this software, the admin can observe and trace the process of data transmission and change some settings to customize activity recognition procedure.

Fig. 3 illustrates all of the processes in our approach and the architecture of the proposed system. On the client side, we have only one module that is *Data Collection*. The client, which is the Data Collector App, receives users' activity data from accelerometer and gyroscope sensors, while true target labels are directly specified by the user. These two data streams are then sent to the server in real-time via the Internet as *Sensors Raw Data* and *User's Feedback*, respectively. All calculations and learning tasks are performed on the server side. First, *Preprocessing Module* receives *Sensors Raw Data* as inputs. It then runs normalization and feature extraction operations to make input data prepared for the recognition phase. The output of this module is a vector which contains normalized feature extracted data (98 features). Once raw input data have been preprocessed, they are then submitted to the *Recognition Module*. This module performs three major tasks, including base model creation, activity detection and model updating. At first, a base model, which is an incremental classifier, is created. This model is trained with the first feature vector produced by the *Preprocessing Module*. At this stage, the model is ready to predict the current activity, but because there is no true label in the system, the first prediction would definitely be incorrect or none. Afterwards, *User's Feedback* containing the true label of current activity comes to the *Recognition Module* in order to first update the base model, and next, to be used for the performance evaluation of the model. This cycle will be repeated for continuously coming data samples to increase prediction accuracy incrementally.

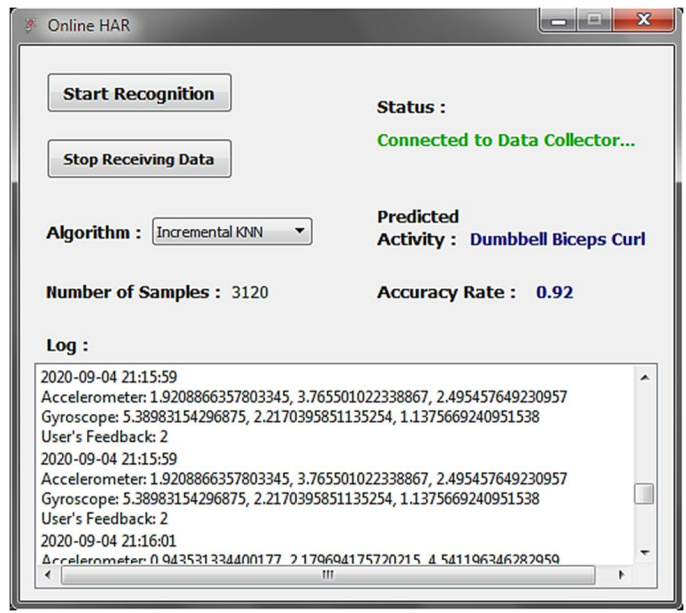

**Fig. 1**. Online HAR monitoring software's UI

## 4. Results and Discussion

In this section, we are going to explore and evaluate the performance of our approach. Experiments are separated into two parts. In the first part, the proposed algorithms are analyzed using evaluation metrics including precision, recall, F1-score and accuracy. As convergence speed is a major determinant of incremental learning methods robustness, we also investigate this factor for all of the presented algorithms using learning curves. In the second part, we compare the performance of our incremental learning algorithms with some of the most prevalent batch learning algorithms as baselines in order to make a meaningful comparison between online and offline HAR solutions. We should mention that for each algorithm, we implemented a large number of models with different configurations to obtain the best results. All the experiments were conducted on a system with an Intel Core2 Duo CPU at 2.10GHz, 4 GB of RAM, 300 GB of secondary storage, running windows 7 and Python 3.7.

### 4.1. Performance Comparison and Convergence Speed

As mentioned in the previous section, we conducted the experiments using four subjects and 20 different activities listed in table 1. To better compare the algorithms, inspired by [35], we first defined a specific scenario for performing the activities by subjects. According to the scenario, each subject picked five activities, which could not be similar to other subjects' activities, and did them this way: 2 minutes of each activity for the first round, then 1 minute of each activity, and another 1 minute of each activity for the final round, which became 20 minutes in total. As data collection frequency is 20Hz, about 24,000 raw sensors data were collected for each subject. The reason behind doing the activities in 3 rounds was to shuffle the data in order to better evaluate the detection power of the algorithms. Due to 2-second sliding windows, after performing feature extraction on the raw data, around 60 data samples in the first round, and 30 data samples in each of the second and third rounds were entered as input data to the algorithms, made a total of 120 data samples for each activity.

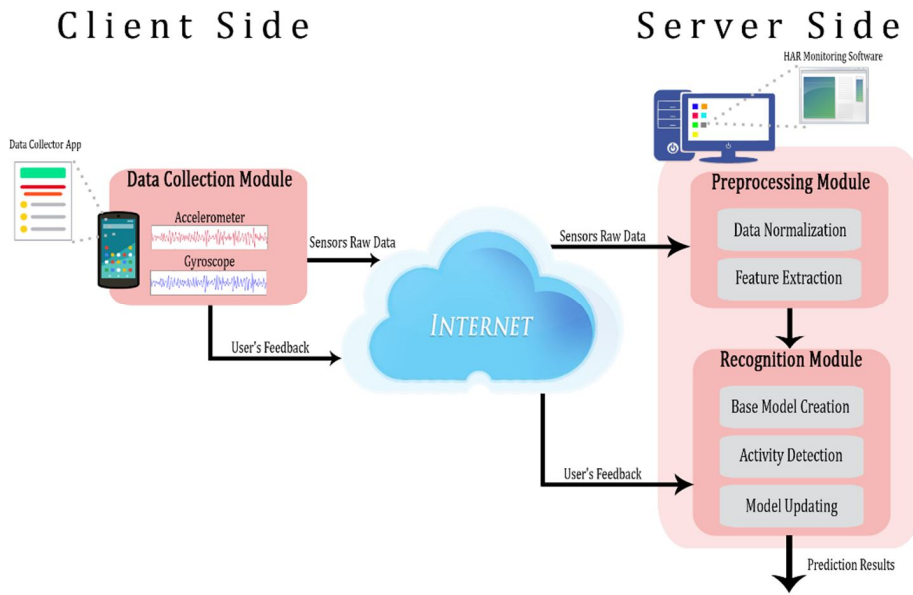

**Fig. 2**. Architecture of online HAR system

**Table 3**. Performance comparison of the algorithms for subject 1

| Algorithms | Precision | Recall | F1-Score | Accuracy | Training Time | Prediction Time |
|---|---|---|---|---|---|---|
| **IKNN** | 96 | 97 | 97 | 96.41 | 0.0013 | 0.1930 |
| **IDT** | 81 | 80 | 80 | 80.71 | 0.0104 | 0.0024 |
| **IRF** | 91 | 91 | 91 | 90.88 | 0.4197 | 0.0063 |
| **IAdaBoost** | 91 | 91 | 91 | 90.73 | 0.0868 | 0.0030 |
| **INB** | 96 | 97 | 97 | 96.41 | 0.0010 | 0.0091 |
| **Learn++NSE** | 91 | 91 | 91 | 90.73 | 0.1350 | 0.0653 |

**Table 4**. Performance comparison of the algorithms for subject 2

| Algorithms | Precision | Recall | F1-Score | Accuracy | Training Time | Prediction Time |
|---|---|---|---|---|---|---|
| **IKNN** | 96 | 96 | 96 | 95.61 | 0.0012 | 0.1726 |
| **IDT** | 73 | 72 | 72 | 71.91 | 0.0101 | 0.0023 |
| **IRF** | 90 | 89 | 89 | 89.12 | 0.4080 | 0.0062 |
| **IAdaBoost** | 85 | 84 | 84 | 84.10 | 0.1476 | 0.0031 |
| **INB** | 96 | 95 | 96 | 95.29 | 0.0010 | 0.0091 |
| **Learn++NSE** | 91 | 91 | 91 | 90.42 | 0.1223 | 0.0596 |

**Table 5**. Performance comparison of the algorithms for subject 3

| Algorithms | Precision | Recall | F1-Score | Accuracy | Training Time | Prediction Time |
|---|---|---|---|---|---|---|
| **IKNN** | 97 | 97 | 97 | 96.66 | 0.0011 | 0.1759 |
| **IDT** | 75 | 76 | 75 | 75.23 | 0.0098 | 0.0023 |
| **IRF** | 92 | 92 | 92 | 91.74 | 0.4146 | 0.0061 |
| **IAdaBoost** | 90 | 89 | 89 | 89.36 | 0.1168 | 0.0031 |
| **INB** | 96 | 96 | 96 | 95.39 | 0.0010 | 0.0090 |
| **Learn++NSE** | 91 | 91 | 91 | 91.11 | 0.1299 | 0.0650 |

**Table 6**. Performance comparison of the algorithms for subject 4

| Algorithms | Precision | Recall | F1-Score | Accuracy | Training Time | Prediction Time |
|---|---|---|---|---|---|---|
| **IKNN** | 98 | 97 | 98 | 97.33 | 0.0012 | 0.1865 |
| **IDT** | 83 | 82 | 83 | 82.26 | 0.0105 | 0.0024 |
| **IRF** | 92 | 92 | 92 | 91.36 | 0.4070 | 0.0064 |
| **IAdaBoost** | 88 | 88 | 88 | 87.59 | 0.1149 | 0.0031 |
| **INB** | 97 | 96 | 97 | 96.23 | 0.0011 | 0.0089 |
| **Learn++NSE** | 91 | 91 | 91 | 90.89 | 0.1408 | 0.0747 |

Tables 3 to 6 show the performance of proposed incremental learning algorithms for different subjects based on evaluation metrics. Since the data is equally distributed between classes (activities), the accuracy rate can well demonstrate the ability of algorithms in recognizing activities. However, for more precise comparisons, we employed other well-known metrics.

Looking at these tables, we can see that IKNN and INB had the best performance, so that the accuracy rate for IKNN was between 95.61% and 97.33% and for INB was in the range of 95.29% to 96.41%, which are excellent results for incremental learning algorithms. There are some reasons showing that KNN and NB benefit from characteristics making their incremental versions powerful for real-time learning. Both of them require small amount of training data for producing great results. They can quickly respond to input data changes and easily update themselves in the arrival of new data samples. They are also less sensitive to outliers and irrelevant values. Moreover, KNN is an instance-based learner and employs local information, making it a highly adaptive algorithm.

We can see that IKNN and INB were followed by IRF and Learn++NSE. These two algorithms, which are from the family of ensemble algorithms, obtained very similar results in almost all cases. However, their best performances were related to subject 3, where IRF and Learn++NSE reached 91.11% and 91.74% accuracy, respectively. The next ensemble algorithm is IAdaBoost, which except for subject 1, its accuracy was below 90% in other cases. It had the weakest performance for subject 2, with an accuracy of no more than 84.10%.

Finally, the worst outcomes belong to IDT that could not compete with its rivals. This algorithm experienced the most considerable fluctuation for different subjects and its classification accuracy was 82.26% at best and 71.91% at worst. These poor results stem from the fact that IDT is more error-prone in classification problems with small number of training data. In fact, when the type of activity changes, IDT needs more input samples for the new activity to complete its learning process. Thus, it is more sensitive to sudden changes of activities which can result in considerable modifications in the tree structure. As in our experiments the number of input instances after feature extraction was not that high, this IDT's requirement did not satisfied. However, keep in mind that more data means more time for processing and learning, which makes our approach far from its major purpose that is real-time activity recognition. Hence, increasing the number of samples more than this seemed unreasonable.

There are two other columns in tables 3 to 6, namely training time and prediction time. It should be mention that processing or execution time is calculated differently for batch and incremental machine learning algorithms. For batch models, the execution time for training stage is equal to the time a learning algorithm spends processing the entire training data. Execution time for test stage can be defined similarly but for test data. However, for incremental learning models, the processing time of only "one" data instance is important and the execution time of an algorithm on the entire dataset is insignificant. The reason is that data samples enter into an incremental learning algorithm one by one, not in the form of batches. Hence, in both prediction (test) and training stages, it is just important how fast the algorithm processes the current input sample. According to these explanations, in tables 3 to 6, the training time column represents the average processing time of one input data sample at the training stage, while the prediction time column shows the average processing time of one input data sample at the prediction stage. Given the figures achieved for all four subjects, in the training stage, INB and IKNN had the shortest processing time and are therefore the fastest algorithms. By contrast, IRF was the slowest one in the training stage, with an average of about 0.4 seconds. In the prediction stage, the situation was reversed and IKNN turned into the slowest algorithm with 0.17-0.19 seconds, and IDT processed each input data faster than its competitors, at about 0.002 seconds. If we consider the aggregation of the training and prediction time, INB and IDT were the fastest algorithms, respectively.

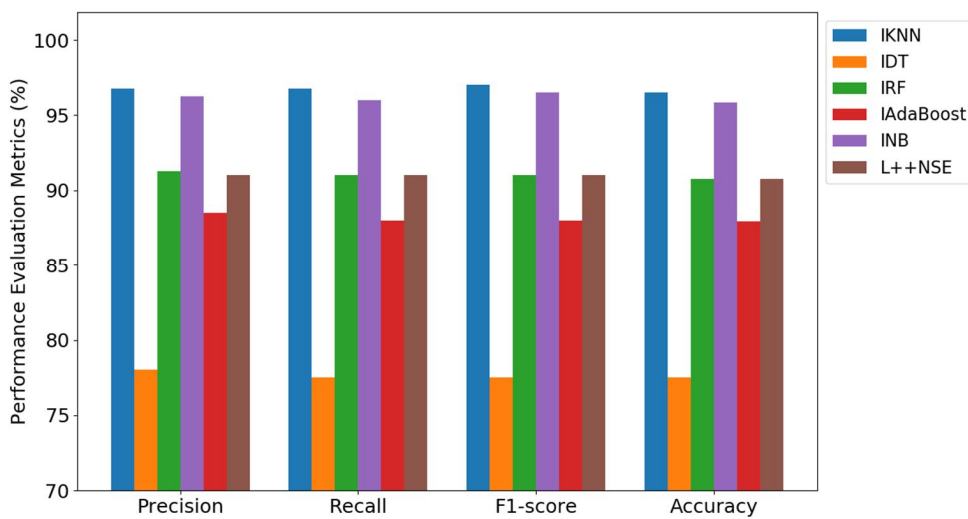

**Fig. 3**. Average results of evaluation metrics of the algorithms for all subjects

Fig. 4 gives a general view of the performance of algorithms, suitable for a thorough comparison. It illustrates the average value of each evaluation metric for all subjects. It can clearly be seen that IKNN acquired the best results for all metrics. INB was the second best algorithm and its difference with IKNN was very small, approximately 1%. Interestingly, despite various values obtained by IRF and Learn++NSE for different subjects, the average results of these two algorithms not only for accuracy rate, but also for other metrics were extremely close, around 91 %. The shortest bars belong to IDT, with 75.5% on average for each metric.

The convergence speed of the algorithms using their learning curves is indicated in fig. 5. Convergence speed indicates the learning speed of algorithms. In other words, it shows changes in the accuracy rate of algorithms when the number of input data samples increases. Each time the type of activity changed, the algorithms experienced a slight decrease in their accuracy rate and then tried to learn and find the pattern of new activity in the shortest possible time. As a

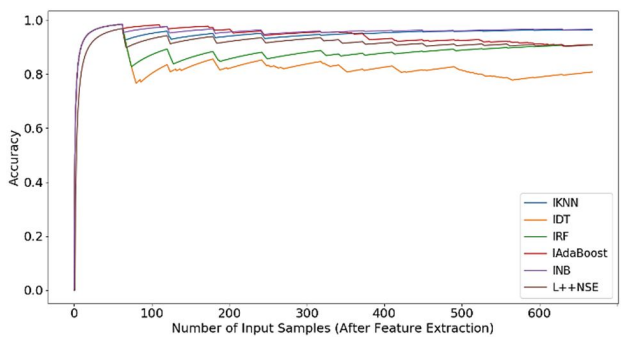

a. Subject 1

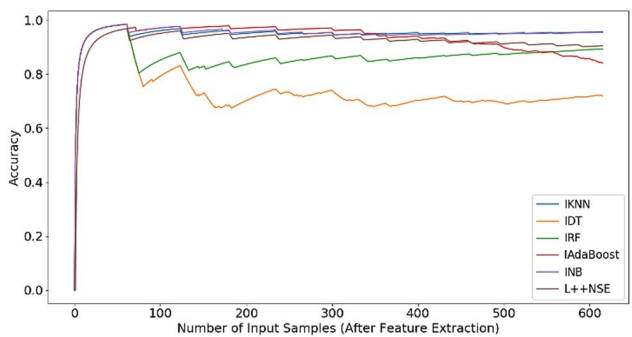

b. Subject 2

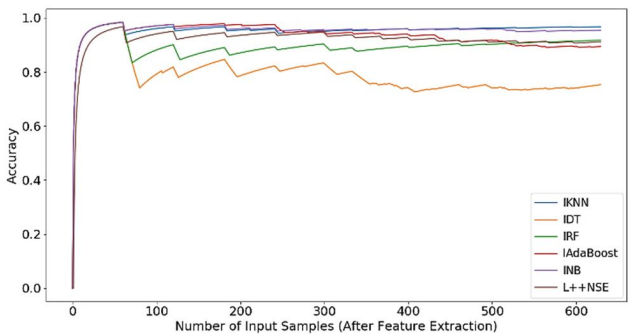

c. Subject 3

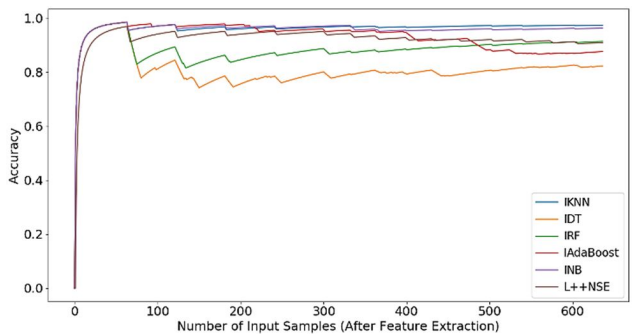

d. Subject 4

**Fig. 5**. Learning curves of the algorithms for different subjects

consequence, the learning curves of algorithms went up after a slight decrease, resulting in the serrated shape of lines. According to fig. 5 (a-d), for all subjects, IKNN and INB had the best convergence speeds and learned faster and finally gained the highest recognition accuracy. A noteworthy point is that each time the current activity changed, IAdaBoost could not adapt well to new input data and its accuracy declines. The faster the activities changed (i.e. from about 300 input samples onwards), the more the accuracy of IAdaBoost reduced, and as we can see, the curve of this algorithm is descending. For Subject 2, the accuracy reduction is the fastest. IRF, on the other hand, followed an opposite trend, trying to recover well after each activity change and learn well from past data. As a result, its learning curve is ascending for all subjects and the changes of activities did not have considerable negative impacts on the learning rate of IRF. Learn++NSE showed an almost constant learning process. That is, after the first reduction in its accuracy rate, it tried to maintain the accuracy at a constant level very close to the final percentage. Finally, there is IDT, which, as mentioned earlier, had the weakest performance. It can be seen that IDT experienced the greatest decrease in recognition accuracy after the first activity change, and even after that, it was unable to regain the lost accuracy.

## 4.2. Online vs. Offline HAR

In this section, based on data obtained from all subjects, we examine the offline HAR system. In fact, we use the offline or batch version of proposed incremental learning algorithms as baselines to compare the performance of our online system with its offline counterpart. Since Learn++NSE does not have an offline version, we made this comparison for the other five algorithms. Considering offline algorithms have sufficient opportunity for learning and are initially trained with a large number of training data, they can obviously perform very well in detecting classes of activities in the test data. Thus, in this section, we are going to use the results acquired from the previous section to discover how much is the gap between the performance of algorithms in batch and incremental mode.

Table 7 shows the performance of baseline algorithms for all subjects. This table provides information about the accuracy of training and test sets along with the F1-score of test set. According to our estimations, batch algorithms achieved great results and never fell below 93% for the two metrics. The training set accuracy of DT and RF reached 100% for all subjects. AdaBoost also gained complete training set accuracy for subjects 3 and 4. In other cases, the classification rates of algorithms for the training set were more than 97%. In relation to test set, the minimum accuracy was about 93%, obtained by DT for subjects 2 and 3. However, in other cases, the recognition accuracy of algorithms for the test set was at least 95%. Overall, the best performance was related to RF whose accuracy for the test set was higher than 96.82% and its F1-score value never fell below 97%. As shown in [61], RF algorithm has always performed exceptionally well for many offline machine learning applications.

Comparing the values of this table with the results obtained from the previous section, we will get some interesting points. First of all, IKNN and INB algorithms performed as good as their batch versions. They achieved an accuracy of 95%-97%, whereas the accuracy rate of KNN and NB for the test set was not much better, between 95% and 98%. This shows that, as we discussed in section 4.1, the incremental versions of these two algorithms have high potential in designing online HAR systems, which further makes them suitable for other online learning applications. The next noteworthy point is the outstanding performance of DT in batch mode, which is completely different from its incremental version performance. While IDT obtained an accuracy of 82.26% and an F1-score of 83% at best, the test set accuracy of DT for subjects 1 and 4 was higher than 98% and its F1-score for subject 4 hit 99%. The other two algorithms, RF and AdaBoost, also showed better performances than their incremental counterparts, especially for subject 2, where RF performed about 9% better than IRF and AdaBoost outperformed IAdaBoost by 14%.

**Table 7**. Performance comparison of offline algorithms for all subjects

| Algorithms | KNN | | DT | | RF | | NB | | AdaBoost | |
|---|---|---|---|---|---|---|---|---|---|---|
| Metrics | Accuracy (Test/Training) | F1-score | Accuracy (Test/Training) | F1-score | Accuracy (Test/Training) | F1-score | Accuracy (Test/Training) | F1-score | Accuracy (Test/Training) | F1-score |
| Subject 1 | 97.01/97.83 | 97 | 98.50/1.00 | 98 | 97.01/1.00 | 97 | 95.52/98.17 | 95 | 95.52/99.00 | 95 |
| Subject 2 | 95.16/98.37 | 95 | 93.54/1.00 | 93 | 98.38/1.00 | 98 | 96.77/98.55 | 97 | 98.38/99.63 | 98 |
| Subject 3 | 95.23/98.41 | 95 | 93.65/1.00 | 94 | 96.82/1.00 | 97 | 96.82/98.58 | 97 | 96.82/1.00 | 97 |
| Subject 4 | 96.87/99.12 | 97 | 98.43/1.00 | 99 | 98.43/1.00 | 99 | 98.43/99.47 | 99 | 98.43/1.00 | 99 |

## 5. Conclusion and Future Work

Offline HAR, which is based on static learning, is not appropriate for real-time prediction of human physical movements, in that the style of activities done by a person can change during the time and the pattern of performing activities can be quite different from person to person. Thus, in this paper, we proposed an online HAR system that is able to detect different activities in real-time without receiving huge amount of data as input for training. To design the system, we used several incremental learning algorithms and evaluated them through comprehensive experiments. According to the results, Incremental K-Nearest Neighbors (IKNN) and Incremental Naïve Bayesian (INB) showed the best performance and could achieve a recognition accuracy of 95-97% for different subjects. Moreover, a comparison between incremental learning algorithms and their batch versions as baselines confirmed the robustness of IKNN and INB in developing online HAR solutions. INB also became the fastest learner. Consequently, we designed our final HAR system using these two algorithms. In addition, we developed a mobile data collector application leveraging accelerometer and gyroscope sensors for sensing activity data and transmitting them to the monitoring system via the Internet.

For future work, we plan to extend our system and develop a light version of online activity recognizer for implementation on users smart devices (smartphone, smart watch, etc.) using incremental learning algorithms. Owing to the limited resources of these devices, some modifications in the structure of HAR approach and algorithms seem necessary to reduce energy consumption, memory usage and computational complexity.

## References

[1] A. Subasi, M. Radhwan, R. Kurdi, and K. Khateeb, "IoT based mobile healthcare system for human activity recognition," in *2018 15th Learning and Technology Conference (L&T)*, 2018, pp. 29-34.

[2] A. Jalal, S. Kamal, and D. Kim, "A Depth Video-based Human Detection and Activity Recognition using Multi-features and Embedded Hidden Markov Models for Health Care Monitoring Systems," *International Journal of Interactive Multimedia & Artificial Intelligence,* vol. 4, no. 4, pp. 54-62, 2017.

[3] J. Ye, X. Li, X. Zhang, Q. Zhang, and W. Chen, "Deep learning-based human activity real-time recognition for pedestrian navigation," *Sensors,* vol. 20, no. 9, p. 2574, 2020.

[4] Y. Xing, C. Lv, H. Wang, D. Cao, E. Velenis, and F.-Y. Wang, "Driver activity recognition for intelligent vehicles: A deep learning approach," *IEEE Transactions on Vehicular Technology,* vol. 68, no. 6, pp. 5379-5390, 2019.

[5] M. Babiker, O. O. Khalifa, K. K. Htike, A. Hassan, and M. Zaharadeen, "Automated daily human activity recognition for video surveillance using neural network," in *2017 IEEE 4th International Conference on Smart Instrumentation, Measurement and Application (ICSIMA)*, 2017, pp. 1-5.

[6] T. Singh and D. K. Vishwakarma, "Human activity recognition in video benchmarks: A survey," in *Advances in Signal Processing and Communication*, Springer, 2019, pp. 247-259.

[7] C. Hu, Y. Chen, L. Hu, and X. Peng, "A novel random forests based class incremental learning method for activity recognition," *Pattern Recognition,* vol. 78, pp. 277-290, 2018.

[8] M. Kose, O. D. Incel, and C. Ersoy, "Online human activity recognition on smart phones," in *Workshop on Mobile Sensing: From Smartphones and Wearables to Big Data*, 2012, pp. 11-15.

[9] C. F. Crispim-Junior, A. Gómez Uría, C. Strumia, M. Koperski, A. König, F. Negin*, et al.*, "Online recognition of daily activities by color-depth sensing and knowledge models," *Sensors,* vol. 17, no. 7, p. 1528, 2017.

[10] M. Janidarmian, A. Roshan Fekr, K. Radecka, and Z. Zilic, "A comprehensive analysis on wearable acceleration sensors in human activity recognition," *Sensors,* vol. 17, no. 3, p. 529, 2017.

[11] Y. Chen and Y. Xue, "A deep learning approach to human activity recognition based on single accelerometer," in *2015 IEEE international conference on systems, man, and cybernetics*, 2015, pp. 1488-1492.

[12] C. Chen, R. Jafari, and N. Kehtarnavaz, "A survey of depth and inertial sensor fusion for human action recognition," *Multimedia Tools and Applications,* vol. 76, no. 3, pp. 4405-4425, 2017.

[13] S. Wang and G. Zhou, "A review on radio based activity recognition," *Digital Communications and Networks,* vol. 1, no. 1, pp. 20-29, 2015.

[14] C. Medrano, I. Plaza, R. Igual, Á. Sánchez, and M. Castro, "The effect of personalization on smartphone-based fall detectors," *Sensors,* vol. 16, no. 1, p. 117, 2016.

[15] J. Suto, S. Oniga, C. Lung, and I. Orha, "Comparison of offline and real-time human activity recognition results using machine learning techniques," *Neural computing and applications,* no. pp. 1-14, 2018.

[16] X. Yin, W. Shen, and X. Wang, "Incremental clustering for human activity detection based on phone sensor data," in *2016 IEEE 20th International Conference on Computer Supported Cooperative Work in Design (CSCWD)*, 2016, pp. 35-40.

[17] J. Morales and D. Akopian, "Physical activity recognition by smartphones, a survey," *Biocybernetics and Biomedical Engineering,* vol. 37, no. 3, pp. 388-400, 2017.

[18] M. Shoaib, S. Bosch, O. D. Incel, H. Scholten, and P. J. Havinga, "A survey of online activity recognition using mobile phones," *Sensors,* vol. 15, no. 1, pp. 2059-2085, 2015.

[19] W. Sousa Lima, E. Souto, K. El-Khatib, R. Jalali, and J. Gama, "Human activity recognition using inertial sensors in a smartphone: An overview," *Sensors,* vol. 19, no. 14, p. 3213, 2019.

[20] O. D. Incel, M. Kose, and C. Ersoy, "A review and taxonomy of activity recognition on mobile phones," *BioNanoScience,* vol. 3, no. 2, pp. 145-171, 2013.

[21] J. Wang, Y. Chen, S. Hao, X. Peng, and L. Hu, "Deep learning for sensor-based activity recognition: A survey," *Pattern Recognition Letters,* vol. 119, pp. 3-11, 2019.

[22] A. Duque, F. J. Ordóñez, P. de Toledo, and A. Sanchis, "Offline and online activity recognition on mobile devices using accelerometer data," in *International Workshop on Ambient Assisted Living*, 2012, pp. 208-215.

[23] H. L. Cardoso and J. M. Moreira, "Human activity recognition by means of online semi-supervised learning," in *2016 17th IEEE International Conference on Mobile Data Management (MDM)*, 2016, pp. 75-77.

[24] S. C. Hoi, D. Sahoo, J. Lu, and P. Zhao, "Online learning: A comprehensive survey," *arXiv preprint arXiv:1802.02871,* 2018.

[25] J. Gama, I. Žliobaitė, A. Bifet, M. Pechenizkiy, and A. Bouchachia, "A survey on concept drift adaptation," *ACM computing surveys (CSUR),* vol. 46, no. 4, pp. 1-37, 2014.

[26] E. Garcia-Ceja and R. Brena, "Building personalized activity recognition models with scarce labeled data based on class similarities," in *International conference on ubiquitous computing and ambient intelligence*, 2015, pp. 265-276.

[27] P. Siirtola, H. Koskimäki, and J. Röning, "Personalizing human activity recognition models using incremental learning," *arXiv preprint arXiv:1905.12628,* no. 2019.

[28] M. L. Gadebe, O. P. Kogeda, and S. O. Ojo, "Personalized Real Time Human Activity Recognition," in *2018 5th International Conference on Soft Computing & Machine Intelligence (ISCMI)*, 2018, pp. 147-154.

[29] X. Sun, H. Kashima, and N. Ueda, "Large-scale personalized human activity recognition using online multitask learning," *IEEE Transactions on Knowledge and Data Engineering,* vol. 25, no. 11, pp. 2551-2563, 2012.

[30] P. Siirtola and J. Röning, "Incremental Learning to Personalize Human Activity Recognition Models: The Importance of Human AI Collaboration," *Sensors,* vol. 19, no. 23, p. 5151, 2019.

[31] A. Gepperth and B. Hammer, "Incremental learning algorithms and applications," in *European Symposium on Artificial Neural Networks (ESANN)*, 2016, pp. 357-368.

[32] M. Chen, S. Mao, and Y. Liu, "Big data: A survey," *Mobile networks and applications,* vol. 19, no. 2, pp. 171-209, 2014.

[33] V. Losing, B. Hammer, and H. Wersing, "Incremental on-line learning: A review and comparison of state of the art algorithms," *Neurocomputing,* vol. 275, pp. 1261-1274, 2018.

[34] L. Mo, Z. Feng, and J. Qian, "Human daily activity recognition with wearable sensors based on incremental learning," in *2016 10th International Conference on Sensing Technology (ICST)*, 2016, pp. 1-5.

[35] Z. Zhao, Z. Chen, Y. Chen, S. Wang, and H. Wang, "A class incremental extreme learning machine for activity recognition," *Cognitive Computation,* vol. 6, no. 3, pp. 423-431, 2014.

[36] N.-Y. Liang, G.-B. Huang, P. Saratchandran, and N. Sundararajan, "A fast and accurate online sequential learning algorithm for feedforward networks," *IEEE Transactions on neural networks,* vol. 17, no. 6, pp. 1411-1423, 2006.

[37] Z. S. Abdallah, M. M. Gaber, B. Srinivasan, and S. Krishnaswamy, "Adaptive mobile activity recognition system with evolving data streams," *Neurocomputing,* vol. 150, pp. 304-317, 2015.

[38] Z. S. Abdallah, M. M. Gaber, B. Srinivasan, and S. Krishnaswamy, "StreamAR: incremental and active learning with evolving sensory data for activity recognition," in *2012 IEEE 24th International Conference on Tools with Artificial Intelligence*, 2012, pp. 1163-1170.

[39] C. Gupta and R. Grossman, "Genic: A single pass generalized incremental algorithm for clustering," in *Proceedings of the 2004 SIAM International Conference on Data Mining*, 2004, pp. 147-153.

[40] A. Gupta, K. Gusain, and D. Kumar, "Novel approach for incremental learning using ensemble of SVMs with particle swarm optimization," in *2016 11th International Conference on Industrial and Information Systems (ICIIS)*, 2016, pp. 426-430.

[41] D.-N. Lu, T.-T. Nguyen, T.-H. Nguyen, and H.-N. Nguyen, "Mobile online activity recognition system based on smartphone sensors," in *International Conference on Advances in Information and Communication Technology*, 2016, pp. 357-366.

[42] A. Duque, F. J. Ordóñez, P. de Toledo, and A. Sanchis, "Offline and online activity recognition on mobile devices using accelerometer data," in *International Workshop on Ambient Assisted Living*, 2012, pp. 208-215.

[43] A. Mannini and S. S. Intille, "Classifier personalization for activity recognition using wrist accelerometers," *IEEE journal of biomedical and health informatics,* vol. 23, no. 4, pp. 1585-1594, 2018.

[44] A. Mannini, M. Rosenberger, W. L. Haskell, A. M. Sabatini, and S. S. Intille, "Activity recognition in youth using single accelerometer placed at wrist or ankle," *Medicine and science in sports and exercise,* vol. 49, no. 4, p. 801, 2017.

[45] B. Krawczyk, "Active and adaptive ensemble learning for online activity recognition from data streams," *Knowledge-Based Systems,* vol. 138, pp. 69-78, 2017.

[46] B. Lika, K. Kolomvatsos, and S. Hadjiefthymiades, "Facing the cold start problem in recommender systems," *Expert Systems with Applications,* vol. 41, no. 4, pp. 2065-2073, 2014.

[47] A. Ignatov, "Real-time human activity recognition from accelerometer data using Convolutional Neural Networks," *Applied Soft Computing,* vol. 62, pp. 915-922, 2018.

[48] M. Christ, N. Braun, J. Neuffer, and A. W. Kempa-Liehr, "Time series feature extraction on basis of scalable hypothesis tests (tsfresh–a python package)," *Neurocomputing,* vol. 307, pp. 72-77, 2018.

[49] M. Barandas, D. Folgado, L. Fernandes, S. Santos, M. Abreu, P. Bota*, et al.*, "TSFEL: Time Series Feature Extraction Library," *SoftwareX,* vol. 11, p. 100456, 2020.

[50] O. Banos, J.-M. Galvez, M. Damas, H. Pomares, and I. Rojas, "Window size impact in human activity recognition," *Sensors,* vol. 14, no. 4, pp. 6474-6499, 2014.

[51] W. Sousa, E. Souto, J. Rodrigres, P. Sadarc, R. Jalali, and K. El-Khatib, "A comparative analysis of the impact of features on human activity recognition with smartphone sensors," in *Proceedings of the 23rd Brazillian Symposium on Multimedia and the Web*, 2017, pp. 397-404.

[52] Z. Yan, V. Subbaraju, D. Chakraborty, A. Misra, and K. Aberer, "Energy-efficient continuous activity recognition on mobile phones: An activity-adaptive approach," in *2012 16th international symposium on wearable computers*, 2012, pp. 17-24.

[53] R. Khusainov, D. Azzi, I. E. Achumba, and S. D. Bersch, "Real-time human ambulation, activity, and physiological monitoring: Taxonomy of issues, techniques, applications, challenges and limitations," *Sensors,* vol. 13, no. 10, pp. 12852-12902, 2013.

[54] R. Elwell and R. Polikar, "Incremental learning of concept drift in nonstationary environments," *IEEE Transactions on Neural Networks,* vol. 22, no. 10, pp. 1517-1531, 2011.

[55] R. Polikar, L. Upda, S. S. Upda, and V. Honavar, "Learn++: An incremental learning algorithm for supervised neural networks," *IEEE transactions on systems, man, and cybernetics, part C (applications and reviews),* vol. 31, no. 4, pp. 497-508, 2001.

[56] P. Domingos and G. Hulten, "Mining high-speed data streams," in *Proceedings of the sixth ACM SIGKDD international conference on Knowledge discovery and data mining*, 2000, pp. 71-80.

[57] A. Saffari, C. Leistner, J. Santner, M. Godec, and H. Bischof, "On-line random forests," in *2009 IEEE 12th international conference on computer vision workshops, iccv workshops*, 2009, pp. 1393-1400.

[58] N. C. Oza, "Online bagging and boosting," in *2005 IEEE international conference on systems, man and cybernetics*, 2005, pp. 2340-2345.

[59] Creme library, Accessed on: Sep. 10, 2021. [Online]. Available: https://github.com/online-ml/river/

[60] J. Montiel, J. Read, A. Bifet, and T. Abdessalem, "Scikit-multiflow: A multi-output streaming framework," *The Journal of Machine Learning Research,* vol. 19, no. 1, pp. 2915-2914, 2018.

[61] M. Vakili, M. Ghamsari, and M. Rezaei, "Performance Analysis and Comparison of Machine and Deep Learning Algorithms for IoT Data Classification," *arXiv preprint arXiv:2001.09636,* 2020.